# Nonlinear Intensity Sonar Image Matching based on Deep Convolution Features

Xiaoteng Zhou[1], Changli Yu[1, *], Xin Yuan[1, *], Yi Wu[2], Haijun Feng[1] and Citong Luo[1]

1. School of Ocean Engineering, Harbin Institute of Technology, Weihai, China, 264209
2. School of Computer Science and Technology, Harbin Institute of Technology, Weihai, China, 264209
Corresponding author: yuchangli@hitwh.edu.cn; xin.yuan@upm.es

*Abstract*—With the continuous development of underwater vision technology, more and more remote sensing images could be obtained. In the underwater scene, sonar sensors are currently the most effective remote perception devices, and the sonar images captured by them could provide rich environment information. In order to analyze a certain scene, we often need to merge the sonar images from different periods, various sonar frequencies and distinctive viewpoints. However, the above scenes will bring nonlinear intensity differences to the sonar images, which will make traditional matching methods almost ineffective. This paper proposes a non-linear intensity sonar image matching method that combines local feature points and deep convolution features. This method has two key advantages: (i) we generate data samples related to local feature points based on the self-learning idea; (ii) we use the convolutional neural network (CNN) and Siamese network architecture to measure the similarity of the local position in the sonar image pair. Our method encapsulates the feature extraction and feature matching stage in a model, and directly learns the mapping function from image patch pairs to matching labels, and achieves matching tasks in a near-end-to-end manner. Feature matching experiments are carried out on the sonar images acquired by autonomous underwater vehicle (AUV) in the real underwater environment. Experiment results show that our method has better matching effects and strong robustness.

## I. Introduction

With the continuous deepening of human development of marine resources, many exploration activities have gradually migrated from shallow seas to deep seas. Sonar is an effective sensor for ocean exploration because it is not interfered by turbidity and medium absorption. Comprehensive exploration tasks often need to combine different platforms and sensors to conduct investigations, and these investigations may be carried out at distinctive operating frequencies, viewpoints and times. Sonar images obtained at different time, viewing angle and sonar type will have tonal differences, which is usually called image nonlinear intensity [1]. Sonar image matching technique usually plays a fundamental role in seabed map stitching, underwater robot navigation and mobile device docking, but the existing image matching algorithms lack robustness for nonlinear intensity difference and have poor matching effect in complex sonar imaging environment [2-4]. Some underwater engineering tasks involving sonar image processing and matching are usually complicated and difficult. Due to the different mechanisms of acoustic imaging and optical imaging, the processing algorithms designed for optical images are often not well used in sonar images. In recent years, CNNs have been widely used in tasks such as image reconstruction, recognition, detection, and segmentation. Its core lies in the ability to extract deep features, which is realized through data-driven [5]. In short, the CNN is equivalent to a high-quality filter. Compared with the classical manually designed filter, it could extract more complex and profound features. Therefore, it is a good tool for sonar images that are tough to design feature filters manually. In this letter, a nonlinear intensity sonar image matching method based on deep convolution features is proposed.

The remainder of this letter is structured as follows. Section II introduces the related work of matching underwater sonar image. Section III details our proposed methodology. Section IV states the setting of our experiment. The evaluation is given in Section V. The conclusions are drawn in Section VI.

## II. Related Work

Since the sonar image matching technology is the basis of the upper-level comprehensive task, many scholars have conducted research on it. In the study of traditional matching methods, King [6] compared the performance of classic matching algorithms on side scan sonar (SSS) images and gave complete experimental data. Matching algorithms include the SIFT [7], ORB [8] and so on. The results show that when there is no large nonlinear intensity difference in the sonar image pair, the matching effect of SIFT is the best. Vandrish [9] compared the matching performances on various SSS images based on SIFT, mutual information maximization and logarithmic polarity cross-correlation, and evaluated it through a series of indicators such as execution time and matching accuracy. The results show that SIFT has better performance. With the development of CNN methods, in the research area of underwater detection research, some researchers attempt to use the CNN to solve the matching problem of sonar images. In [10], the author proposed to use the CNN to build a specific similarity evaluation model to solve the matching problems of the forward-looking sonar (FLS) images. This study collected a certain number of FLS image datasets for training and testing. The results show that the matching effects using the CNN is better than that of the classical matching algorithms, such as the SIFT. This research is a successful attempt to introduce the CNN algorithm into the underwater sonar image matching task. The author in [11] to use the CNN network to establish a similarity evaluation model to solve the matching problem of the SSS image, and ultimately serve the AUV autonomous navigation.

In summary, the traditional matching methods are manually designed based on the optical images, they could clearly detect the position of image feature points and they are effective on sonar images with continuous frames, but they are almost unable to complete the matching task when encountering image pairs with nonlinear intensity differences. By contrast, the current CNN-based matching methods could use the deep convolution features to estimate the similarity of the sonar image patches. However, there is no direct correlation between the estimated outputs and the local feature points, and the transformation information of the two images cannot be obtained through matching, and furthermore, it cannot directly serve the tasks such as positioning or mapping in the later stage. Taking into account that the location of local feature points is relatively accurate, and they could ensure the accuracy of the upper-level tasks. This paper proposes a matching method combining the local feature points and deep convolution features to solve the matching problem of nonlinear intensity sonar images.

## III. DETAILED METHODOLOGY

Due to the scarcity of underwater sonar images, and the sonar images in different sea areas are quite different, so we introduce the self-learning idea to construct data samples and train models. Our method is mainly divided into four steps: model input, sample construction, model training, and model prediction. The overall pipeline of our method is depicted in Fig. 1.

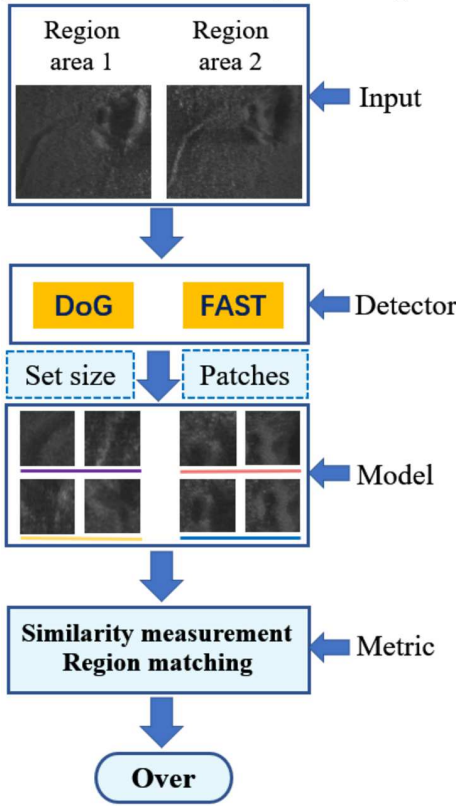

Fig. 1. The pipeline of our matching model.

Firstly, a set of feature detectors are used to detect feature points of two input images that will be used for matching, and map the effect in the image with more feature points to another image. Secondly, with each feature point detected above as the center and a certain preset size as the edge distance, a rectangular partial area is constructed as an image patch, and the image patches are made into positive and negative data samples for subsequent model training. Thirdly, a model centered on the Siamese network architecture is used, and data samples are used for training to complete similarity evaluation. Fourthly, the similarity result of the image region is mapped to each detected feature point for subsequent matching degree judgment and false matching elimination.

### A. Feature Detection

We use the DoG and FAST operators to complete the feature detection. They are used in the common SIFT algorithm and ORB algorithm, respectively, and have good performances.

The performance of the DoG detector is pretty robust, and its features remain unchanged under conditions such as scale conversion, rotation, and lighting. Its shortcomings are that it is not sensitive to edge point features and its speed is relatively slow. The FAST detector only uses surrounding pixels for comparison, and could perform high-speed feature detection at a real-time frame rate. Its disadvantage lies in the lack of feature richness. In this case, we reasonably combine the two detectors to enhance the effects of feature detection.

Before using the combined feature detector for feature detection, we use the sonar echo data and positioning assistance data to strictly align the region area through geocoding. Next, perform feature detection on the two images, separately. Due to the serious nonlinear intensity differences between the two images, the detectors cannot detect rich and uniform feature points. We propose a cross-mapping method to extract rich feature points. First of all, perform stepwise feature detection on two images respectively, and then map the detection effect with more feature points to the opposite image, and finally fuse the raw feature points. The distribution of the proposed method is shown in Fig. 2.

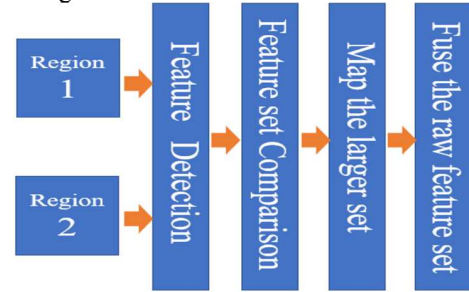

Fig. 2. The pipeline of our proposed feature extraction method.

We take the detection process of the DoG detector as an example to describe our proposed feature point set construction method, and the effect diagram is displayed in Fig. 3. The image in the first row is the raw feature detection effects of the DoG detector. It could be seen that the feature distribution directions of the same geographic area A and B are different, which makes the traditional description process almost invalid. The image in the second row displays the effect of mapping the rich feature detection effect of the image on the right to the left, while the image in the third row is just the opposite. Finally, the feature effects of the two-step detection are integrated, and the effect is shown in the fourth row of the image. At this time, the feature distribution is relatively dense and uniform.

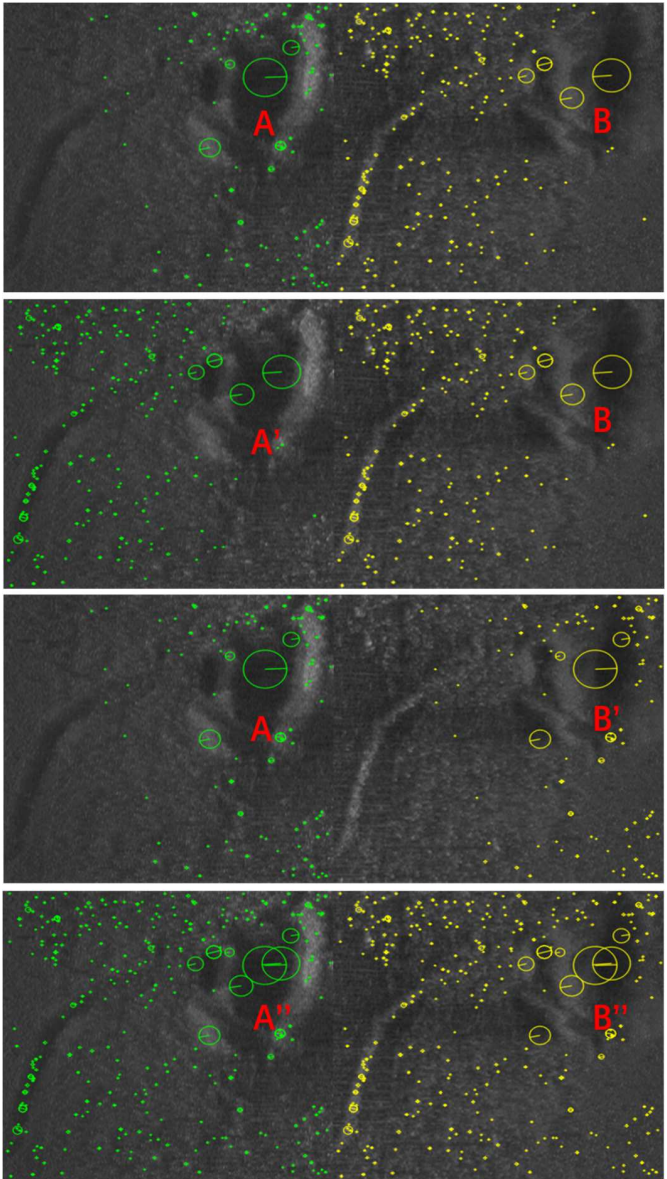

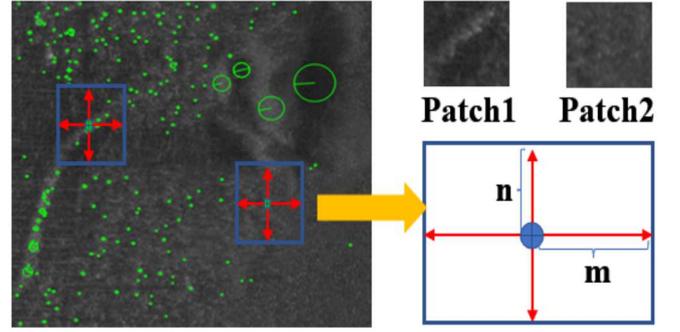

Fig. 4. The schematic diagram of image patches generation, where $m$ and $n$ represent the custom edge distance, they jointly determine the size of the sonar image patches.

### B. Local Similarity Descriptor Construction

We try to use a deep learning framework to directly learn the relationship between image pairs and matching labels. The input of the model network is two image pairs, and the output is its corresponding matching label. Then, the trained deep neural network is used to predict the matching label of the image pair composed of the two images to be registered. The basic Siamese network architecture adopted is shown in Fig. 5.

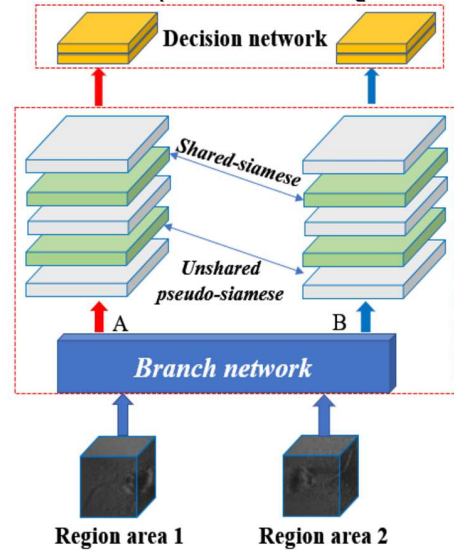

Fig. 3. The schematic diagram of our proposed feature detection method, A and B represents the corresponding geographic area. $A'$ and $B'$ represents the effects after the mapping operation, $A''$ and $B''$ represents the effects of integrating all the feature points.

As can be seen from the first line of the above figure, the features normally detected by the detector do not correspond completely and the sparsity gap is very large, which cannot be used in the subsequent feature description and matching links. However, after processing by our method, the sonar target area has abundant and strictly corresponding feature points, which will help us to design descriptors for them. We perform the same processing on the effects detected by the FAST detector and map them to the DoG effects. Next, we take the feature point to the center, and use $m$, $n$ distance as the edge distance to divide the image patches to construct samples for subsequent model training. Therefore, the positions of the corresponding two points could form one positive sample, indicating that they are matched, and then randomly shuffle one of the image patches to indicate that it is a mismatched pair, that is, a negative sample. The schematic diagram of image patches generation is illustrated in Fig. 4.

Fig. 5. The architecture of the basic Siamese network.

To determine the region area 1 and region area 2 in the sonar image, firstly we need to construct a network mapping function $G_W(X)$, and then use region area 1 and area 2 as the parameter independent variables $X_1$, $X_2$, we could get $G_W(X_1)$, $G_W(X_2)$, and that is, the feature vector used to evaluate whether $X_1$ and $X_2$ are similar is obtained. Next, construct the $Loss$ as follows:

$$E_W(X_1, X_2) = \|G_W(X_1) - G_W(X_2)\| \tag{1}$$

The two-branch weights of the above Siamese network architecture are shared, and then the CNN is used to extract region area features. The top decision layer is used to output the similarity of features. This idea was successfully tried in [12], and the main design of our model is also referred to it. We convert it to a binary classification problem, that is, the output result is 1 for matching, and 0 for non-matching. The detailed architecture of the model is displayed in Fig. 6.

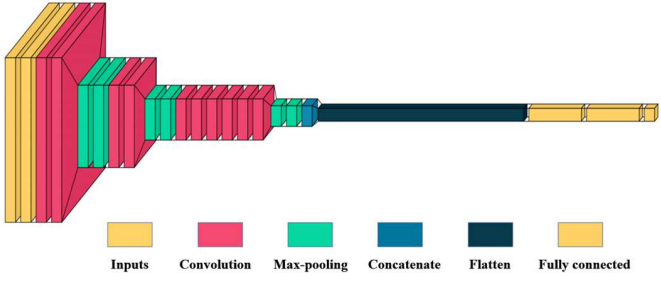

Fig. 6. The detailed schematic of our model.

*C. Feature Matching*

For the traditional matching methods, there is no information feedback between feature extraction and feature matching, so feature extraction can not be adjusted adaptively according to the image to be matched. We try to extract local features, and then use the deep convolution network to further design the descriptors of these features, so that they could be combined effectively to perform the subsequent matching task. Feature extraction and feature matching are unified in an end-to-end framework, and the mapping function from image patch pairs to matching labels is directly learned.

During the training process, we have no requirement on the size of each input patch, and we could freely combine different size such as 16x16, 32x32 or 64x64, to achieve better matching performances. On the basis of the above samples, we introduced the data enhancement and transfer learning ideas to improve the generalization ability of the model while reducing the impact of overfitting. Data enhancement operations include adding noise, rotation, translation, and scale transformation. The detail of transfer learning is to use the complete sonar waterfall map on the both sides of Fig. 7 for pretraining to improve the ability of the model to extract features within the sonar images.

We map the local similarity evaluation results predicted by the model to local key points to determine whether it matches. If the model prediction probability is close to 1 and greater than the setting similarity estimation threshold, the output is 1, which means the two images do match. On the contrary, the model output is 0, which implies that they do not match, and finally the entire task is transformed into a binary classification task.

## IV. EXPERIMENT

The SSS images on the Fig. 7 are obtained by Deep Vision AB company [13] using the DeepEye 680D in Lake Vättern, Sweden. We select a group of nonlinear intensity sonar image regions for the subsequent matching test, and the amplified intensity differences are shown in Fig. 8.

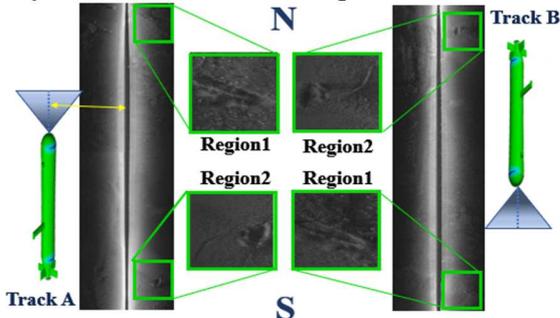

Fig. 7. The repeated detection path of the AUV carrying sonar from South to North, and then from North to South.

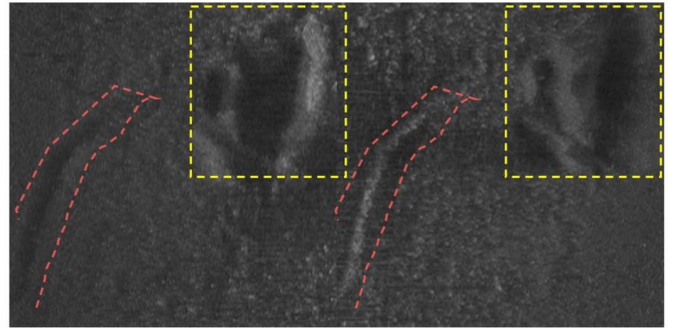

Fig. 8. A pair of SSS images with nonlinear intensity differences.

It could be seen from the feature details of the sonar images shown in Fig. 8 that when the same target area is detected from different directions, there are obvious differences in the edge contour and gray display of the image. In this case, the manually designed local feature filter can not maintain good performance. However, in the actual exploration process, the sonar images with nonlinear intensity differences are the mainstream, which makes the sonar image matching task facing great challenges.

## V. EXPERIMENTAL EVALUATION

We compared the overall matching effects of the classic and state-of-the-art approaches on the sonar images with nonlinear intensity differences. These approaches include the classic image matching methods SIFT, ORB, BRISK [14] and the deep learning-based method SuperPoint [15] and the transformer-based method LoFTR [16]. These methods have shown good performances in many scene matching tasks, and the overall matching effects comparison diagram is shown in Fig. 9.

All methods were implemented under the Windows 10 operating system using Python 3.7 with an Intel Core i7-9700 3.00GHz processor, 16GB of physical memory, and one NVIDIA GeForce RTX2070s graphics card. The SIFT, ORB and BRISK matching approaches are implemented based on the OpenCV-Python tools [17]. In order to maximize the matching performances of the above methods, we have adopted their original parameter settings, in which the matching distance threshold ($d_{ratio}$) of the SIFT, ORB and BRISK is set to 0.85 and the matching mode is the K-Nearest Neighbor (KNN).

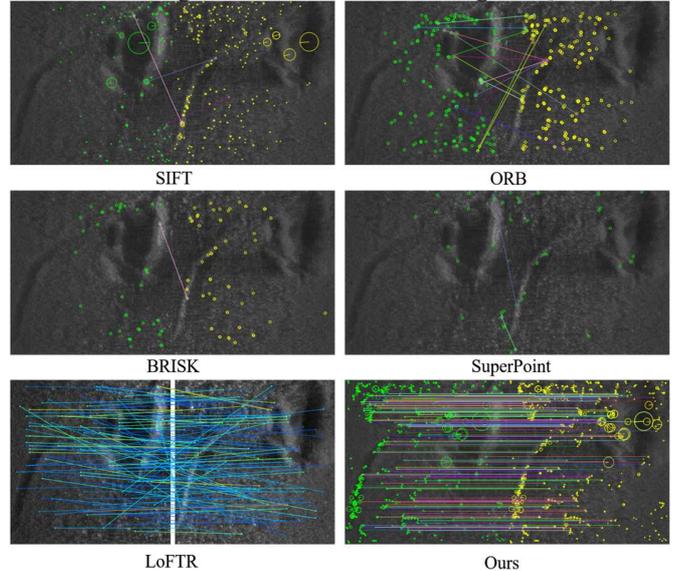

Fig. 9. The overall matching effects comparison diagram.

It could be seen from the overall matching effects that other approaches do not complete the matching task effectively, while our method could achieve accurate matching of local feature points. Since other methods do not achieve matching, this letter does not compare time and other performances. In subsequent tasks, we will apply the model to a richer sonar image dataset, such as rotation and scale transformation, instead of limited to the translation state to improve its adaptability. The construction method of sonar sample data set based on self-learning idea proposed by us could provide a reference for the design of sonar image matching algorithms based on deep learning in the future.

## VI. Conclusion

This letter proposes a nonlinear intensity sonar image matching method based on local feature points and deep convolution features. Our method uses a model with a Siamese network architecture as the core to extract deep convolutional features of the object region and evaluate their similarity. Next, we associate the evaluation results with the locations of local feature points in the region to complete the matching task. The matching tests are carried out on the real SSS sonar images, and the results prove that it could better deal with the problem of nonlinear intensity changes between sonar image pairs. In the future, we will collect more sonar images to expand the sample data set, and further expand the self-learning method to enrich the dimension of data features. In the matching stage, we will try more advanced feature detectors to improve the overall matching performances, and to replace the feature extraction network in the Siamese architecture with other networks with stronger extraction capabilities.

## Acknowledgment

The research was undertaken in the School of Ocean Engineering, Harbin Institute of Technology and the authors also would like to acknowledge the work of Yi Wu in the development of the image preprocessing software. At the same time, thanks to the sonar image data support provided by Deep Vision AB company.

## References

[1] J. E. H. Clarke, K. K. Iwanowska, D. R. Parrott, G. P. Duffy, M. Lamplugh and J. Griffin, "Inter-calibrating multi-source, multi-platform backscatter data sets to assist in compiling regional sediment type maps: Bay of Fundy," in the Canadian Hydrographic Conference and National Surveyors Conference, 2008.
[2] M. Lacharité, C. J. Brown, and V. Gazzola, "Multisource multibeam backscatter data: developing a strategy for the production of benthic habitat maps using semi-automated seafloor classification methods," Marine Geophysical Research, vol. 39, pp. 307–322, 2018.
[3] X. Jiang, J. Ma, G. Xiao, Z. Shao and X Guo, "A review of multimodal image matching: Methods and applications," Information Fusion, vol. 73, pp. 22-71, 2021.
[4] X. Shang, J. Zhao, and H. Zhang, "Automatic Overlapping Area Determination and Segmentation for Multiple Side Scan Sonar Images Mosaic," IEEE Journal of Selected Topics in Applied Earth Observations and Remote Sensing, vol. 14, pp. 2886-2900, 2021.
[5] D. Bhatt, C. Patel, H. Talsania, J. Patel, R. Vaghela and S. Pandya, "CNN Variants for Computer Vision: History, Architecture, Application, Challenges and Future Scope," Electronics, vol. 10, no. 20, p. 2470, 2021. [Online]. Available: https://www.mdpi.com/2079-9292/10/20/2470.
[6] P. King, B. Anstey, and A. Vardy, "Comparison of feature detection techniques for AUV navigation along a trained route, " 2013 OCEANS - San Diego, pp. 1-8, 2013.
[7] D. G. Lowe, "Distinctive Image Features from Scale-Invariant Keypoints," International Journal of Computer Vision, vol. 60, no. 2, pp. 91-110, 2004.
[8] E. Rublee, V. Rabaud, K. Konolige, and G. Bradski, "ORB: An efficient alternative to SIFT or SURF," in 2011 International Conference on Computer Vision, pp. 2564-2571, 2011.
[9] P. Vandrish, A. Vardy, D. Walker, and O. A. Dobre, "Side-scan sonar image registration for AUV navigation, " in 2011 IEEE Symposium on Underwater Technology and Workshop on Scientific Use of Submarine Cables and Related Technologies, pp. 1-7,2011.
[10] Valdenegro-Toro, and Matias, "Improving Sonar Image Patch Matching via Deep Learning," in 2017 European Conference on Mobile Robots (ECMR), pp. 1-6, 2017.
[11] W. Yang, S. Fan, S. Xu, P. King, B. Kang, and E. Kim, "Autonomous Underwater Vehicle Navigation Using Sonar Image Matching based on Convolutional Neural Network," IFAC-PapersOnLine, vol. 52, no. 21, pp. 156-162, 2019.
[12] X. Han, T. Leung, Y. Jia, R. Sukthankar, and A. C. Berg, "MatchNet: Unifying feature and metric learning for patch-based matching," in Computer Vision & Pattern Recognition, 2015.
[13] Deep Vision AB company. Available online: http://deepvision.se/ (accessed on 15 December 2021).
[14] S. Leutenegger, M. Chli, and R. Y. Siegwart, "BRISK: Binary Robust invariant scalable keypoints," in International Conference on Computer Vision, pp. 2548-2555, 2011.
[15] D. Detone, T. Malisiewicz, and A. Rabinovich, "SuperPoint: Self-Supervised Interest Point Detection and Description," in 2018 IEEE/CVF Conference on Computer Vision and Pattern Recognition Workshops (CVPRW), pp. 337-33712, 2018.
[16] J. Sun, Z. Shen, Y. Wang, H. Bao, and X. Zhou, "LoFTR: Detector-Free Local Feature Matching with Transformers," 2021.
[17] G. Bradski, "The OpenCV Library," dr dobbs journal of software tools, 2000. Available online: http://www.drdobbs.com/open-source/the-opencv-library/184404319 (accessed on 15 December 2021).